\documentclass{article}

\usepackage{PRIMEarxiv}

\usepackage[utf8]{inputenc} 
\usepackage[T1]{fontenc}    
\usepackage{hyperref}       
\usepackage{url}            
\usepackage{booktabs}       
\usepackage{amsfonts}       
\usepackage{nicefrac}       
\usepackage{microtype}      
\usepackage{lipsum}
\usepackage{graphicx}       
\graphicspath{{media/}}     

\usepackage{authblk}

\title{Reasoning before Comparison: LLM-Enhanced Semantic Similarity Metrics for Domain Specialized Text Analysis}

\author[1]{Shaochen Xu}
\author[1]{Zihao Wu}
\author[1]{Huaqin Zhao}
\author[1]{Peng Shu}
\author[1]{Zhengliang Liu}
\author[2]{Wenxiong Liao} 
\author[3]{\\Sheng Li}
\author[4]{Andrea Sikora}
\author[1]{Tianming Liu}
\author[5]{Xiang Li}

\affil[1]{School of Computing, University of Georgia}
\affil[2]{School of Computer Science and Engineering, South China University of Technology}
\affil[3]{School of Data Science, University of Virginia}
\affil[4]{Department of Clinical and Administrative Pharmacy, University of Georgia College of Pharmacy}
\affil[5]{Massachusetts General Hospital and Harvard Medical School}

\begin{document}

\maketitle

\begin{abstract}
In this study, we leverage LLM to enhance the semantic analysis and develop similarity metrics for texts, addressing the limitations of traditional unsupervised NLP metrics like ROUGE and BLEU. We develop a framework where LLMs such as GPT-4 are employed for zero-shot text identification and label generation for radiology reports, where the labels are then used as measurements for text similarity. By testing the proposed framework on the MIMIC data, we find that GPT-4 generated labels can significantly improve the semantic similarity assessment, with scores more closely aligned with clinical ground truth than traditional NLP metrics. Our work demonstrates the possibility of conducting semantic analysis of the text data using semi-quantitative reasoning results by the LLMs for highly specialized domains. While the framework is implemented for radiology report similarity analysis, its concept can be extended to other specialized domains as well. 
    
\end{abstract}

\section{Introduction}
The analysis of medical texts is a key component of healthcare informatics, where the accurate comparison and interpretation of documents can significantly impact patient care and medical research. Traditionally, this analysis has leveraged lexical comparison metrics such as ROUGE (Recall-Oriented Understudy for Gisting Evaluation) \cite{lin-2004-rouge} and BLEU (Bilingual Evaluation Understudy) \cite{papineni-etal-2002-bleu}, which have become standard tools in the evaluation of text similarity within the domain of natural language processing (NLP). ROUGE and BLEU were initially designed to assess the quality of automatic summarization and machine translation respectively, by measuring the overlap of n-grams between the generated texts and reference texts.

While these metrics have been instrumental in advancing NLP applications, their application in medical text analysis reveals inherent limitations. Specifically, ROUGE and BLEU focus predominantly on surface-level lexical similarities, often overlooking the deep semantic meanings and clinical implications embedded within medical documents. This gap in capturing the essence and context of medical language presents a significant challenge in leveraging these metrics for meaningful analysis in healthcare.

Recognizing these limitations, this research proposes a novel methodology that employs GPT-4, a state-of-the-art large language model, for a more sophisticated analysis of medical texts. GPT-4's advanced understanding of context and semantics \cite{liu2023geval, naismith2023automated, fu2023gptscore} offers an opportunity to transcend the boundaries of traditional lexical analysis, enabling a deeper, more meaningful comparison of medical documents \cite{brown2020language, devlin2019bert}. This approach not only addresses the shortcomings of ROUGE and BLEU but also aligns with the evolving needs of medical data analysis, where the accurate interpretation of texts is preeminent.

Despite the proliferation of advanced language models in the general domain, with models like Google's Gemini \cite{geminiteam2023gemini} claiming superior performance to GPT-4 through comprehensive benchmark comparisons, the specialized domain of medical text analysis presents unique challenges that remain largely unaddressed. The intricacies of medical language, characterized by its dense terminologies, expressions, and implications, necessitate models that are not only adept at general language understanding but are also finely tuned to grasp the subtleties of medical discourse. Unlike the broader NLP field where benchmarking new models against established standards like GPT-4 is common, the medical field lacks a similar benchmarking framework to rigorously evaluate the performance of specialized LLMs. This gap highlights a need for developing methodologies that can effectively compare the language generated by AI models directly to ground truth data within medical contexts. Therefore, this study aims to establish a methodological framework that bridges this gap, offering a way to directly compare generated language to the authentic text data found in medical documents. Such a framework would not only enhance our understanding of how well current LLMs perform in specialized domains but also set a precedent for evaluating future models in the context of healthcare informatics, contributing significantly to the advancement of precision medicine and evidence-based clinical practices

An essential aspect of our methodology is the integration of a "human-in-the-loop" approach for evaluating the results. This framework enhances the interpretability of AI-generated outputs, making it easier for medical professionals to engage with and derive insights from complex data analyses. By converting GPT-4’s analysis into generic labels for categorization, we present a more accessible and interpretable medium for comparison. This innovation significantly simplifies the task for human reviewers, who can now assess the similarity and relevance of documents through label comparisons rather than exhaustive text reviews, aligning with the principles of explainable AI (XAI) and enhancing user trust and understanding in AI applications \cite{arrieta2019explainable, holzinger2016interactive}.

The significance of this study extends beyond addressing the limitations of existing lexical metrics; it proposes a multi-dimensional, quantifiable approach for text comparison that meets the granular and diverse needs of the medical field. By integrating the computational strengths of GPT-4 with the insights of human expertise, this research contributes to the advancement of precision medicine and evidence-based practice, promising to revolutionize medical data analysis and healthcare informatics.

\section{Related Works}
\subsection{Text Comparison Methods in Medical Text Analysis}

In healthcare environments, imaging examinations including ultrasonography, magnetic resonance imaging (MRI), and computed tomography (CT) scans, are pivotal for patient assessment and care. To enhance the accuracy and quality of imaging diagnostics, it is essential to meticulously review and contrast pre-existing reports of imaging and pathology examinations that pertain to identical anatomical sites, as recorded in electronic medical records (EMRs). This retrieval process, however, is notably labor-intensive. Consequently, scholars in the field have dedicated efforts to develop a variety of models and metrics aimed at quantifying text similarities, facilitating a more efficient and systematic approach to comparing diagnostic reports\cite{reiter2018structured,gulden2019extractive,lin-2004-rouge,banerjee2005meteor,puppala2015meteor,soualmia2012matching,rahutomo2012semantic,huang2016supervised}.

Researchers have defined numerous metrics to measure medical text similarity, encompassing perspectives from traditional linguistics, and syntactic angles, to similarities in text vector representations. The Bilingual Evaluation Understudy (BLEU) metric, while primarily used for evaluating machine translation quality, can also be applied in comparing medical texts generated from different sources or translation systems, particularly for assessing translation accuracy\cite{reiter2018structured}. The Recall-Oriented Understudy for Gisting Evaluation (ROUGE) metric is extensively utilized in automatic text summarization, especially for creating summaries of medical research reports or patient records, aiding in evaluating the consistency between auto-generated summaries and human-written counterparts\cite{gulden2019extractive,lin-2004-rouge}. Moreover, the Metric for Evaluation of Translation with Explicit Ordering (METEOR) provides a more precise assessment in medical text comparisons, such as patient record descriptions versus medical database entries, by considering synonyms and sentence structures, catering to tasks requiring high linguistic understanding\cite{banerjee2005meteor,puppala2015meteor}. Levenshtein distance is employed to assess and correct spelling errors in medical texts, like patient record entry or typographical errors in medical literature, enhancing accuracy in text comparisons\cite{soualmia2012matching,banerjee2005meteor}. Cosine similarity is used in medical literature retrieval and similar case finding, comparing document vector representations to identify thematically or content-wise similar documents, valuable for researchers and clinicians seeking related studies or cases\cite{ye2015improved,xia2015learning,rahutomo2012semantic}. Lastly, the Word Mover's Distance (WMD), is another metric that captures deep semantic information and proves especially valuable in comparing medical texts, such as clinical case report similarities, identifying semantically similar but differently expressed terms, crucial for handling the rich and varied expressions in medical terminology\cite{huang2016supervised,kusner2015word}.

Furthermore, there is more research exploring potential combination applications between LLMs and healthcare\cite{gong2023evaluating,shi2023mededit,kim2023medivista,liu2023radonc,liu2023artificial}. \cite{liu2023holistic} aims to evaluate GPT-4V across a diverse set of capabilities required for real-world medical imaging tasks in various modalities such as X-ray, MRI, CT, and microscopy images. \cite{zhao2023ophtha} develops an LLM specifically for ophthalmology, called Ophtha-LLaMA2 which exhibits satisfying accuracy and efficiency in ophthalmic disease diagnosis. \cite{zhong2023chatradio}  proposes the ChatRadio-Valuer, a tailored model for automatic radiology report generation that learns generalizable representations and provides a basis pattern for model adaptation in sophisticated analysts’ cases.

\section{Method}
\begin{figure*}[h]
\centering
\includegraphics[width=\textwidth]{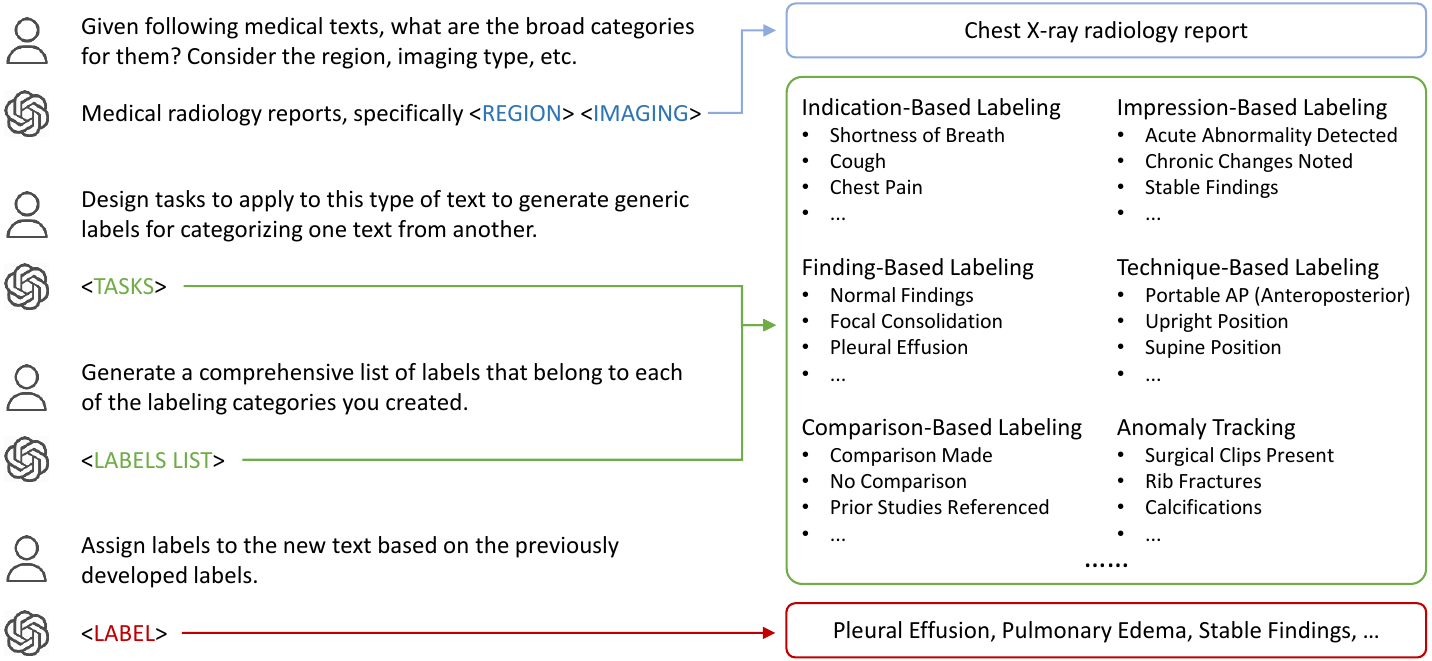}
\caption{Overview of the prompt and response conversation between the user and GPT-4 to generate a list of custom labels for any given medical text. (1) The left side showcases the entire conversational flow between the user and GPT-4 to generate the desired labels. (2) While the right side displays the products of said conversation from GPT-4's response.}
\label{prompt}
\end{figure*}

In this study, we aim to leverage GPT-4 for the comparative analysis of medical texts at a higher semantic level. This method separates from conventional lexical comparison tools, relying on the LLM's ability to interpret, reason, and categorize medical texts based on their content (See Figure \ref{prompt}).

\subsection{Text Identification}
The initial step of the proposed framework focuses on the task of text identification, where we present GPT-4 with anonymized radiology reports, without providing any contextual clues or background information. This step is essential, as it assesses GPT-4's inherent capability to accurately discern the type and nature of the given documents based solely on textual content. The ability of GPT-4 to perform this task is not taken for granted; rather, drawing on the work by Avati et al. \cite{avati2018improving}, who highlighted the role of precise data categorization in healthcare data analysis, our study explores the applicability of LLMs for similar tasks in medical informatics.

The significance of this stage lies in its potential to validate the utility of LLMs in the field of medical informatics, where precision and context are predominant. By successfully identifying medical texts, GPT-4 would demonstrate a critical level of contextual interpretation, which goes beyond basic keyword recognition or surface-level analysis. This step, therefore, sets the stage for more complex analytical tasks, such as the generation of identification tasks and labeling, which are predicated on the model's initial identification accuracy.

\subsection{Task Generation}
Following the identification of medical texts by GPT-4, the next step in our methodology is the generation of tasks by the model. This step involves asking GPT-4 to lay out the specific criteria or features utilized for recognizing and classifying the medical texts. This approach aligns with the principles of explainable AI (XAI) \cite{arrieta2019explainable}, which emphasizes making AI's decision-making processes transparent and understandable to humans. In their work, Arrieta et al. underscores the importance of XAI in enhancing the interpretability of AI models, thereby making their operations and decisions more accessible to users.

The tasks generated by GPT-4 are expected to cover a broad spectrum, reflecting the model's understanding of medical texts. These tasks might include identifying key medical terminologies and their relevance to certain conditions or recognizing the structure of the reports and the implications of specific diagnostic criteria. Importantly, these tasks can offer tailored utility to different medical professionals. For instance, tasks related to detailed findings and diagnostic interpretations could be particularly valuable to radiologists, who require a deep understanding of imaging reports to make accurate diagnoses. On the other hand, tasks that assess the urgency or emergency level of cases, based on textual cues within the reports, could be crucial for nurses and emergency department staff. This differentiation in task relevance underscores the potential of GPT-4 to support a wide range of clinical activities and decision-making processes.

This aspect of task generation draws inspiration from the work of Holzinger et al. \cite{holzinger2016interactive}, who advocate for interactive machine learning (iML) in enhancing the collaboration between humans and AI, especially in complex domains like healthcare. iML focuses on leveraging human expertise to improve AI models' learning and interpretability. By involving "human-in-the-loop" throughout different stages and generating tasks that cater to the specific needs and expertise of different medical professionals, GPT-4 could significantly contribute to the development of AI tools that are not only advanced in their language understanding capabilities but also versatile in their application across the healthcare sector.

\subsection{Label Creation}
The label creation phase, following task generation by GPT-4, is pivotal for organizing the AI-identified features of medical texts into coherent and actionable labels. This process is not just about leveraging GPT-4's advanced language capabilities but also about enhancing the labels' usability and interpretability in clinical settings. By categorizing complex medical information into concise labels, we facilitate a more intuitive interface for medical professionals to engage with AI-generated insights.

An integral aspect of this phase is the incorporation of the "human-in-the-loop" methodology, which significantly enriches the label creation process. This approach allows for the direct involvement of medical professionals in the review and refinement of the generated labels. The practical advantage of this is twofold. First, it makes it easier for human reviewers to discern differences and similarities between sets of labels, rather than navigating the complexities of full medical texts. This comparative simplicity can be crucial in clinical settings, where efficiency and accuracy are imperative. Second, involving human expertise at this stage ensures the labels are not only accurate but also clinically relevant and aligned with medical terminologies and practices. The generated labels thus serve as a bridge between GPT-4's computational analysis and the practical knowledge and needs of healthcare practitioners. Such an approach aligns with the work of Caruana et al. \cite{caruana2015intelligible}, who emphasized the importance of interpretability in medical AI applications, suggesting that models should be developed with a focus on their eventual use in patient care.

Furthermore, the systematic categorization provided by these labels supports the broader goal of integrating AI tools into healthcare workflows in a manner that enhances, rather than complicates, decision-making processes. By ensuring that labels are both generated and validated with human insight, the label creation phase becomes a fundamental step towards developing AI applications that are not only powerful in analysis but also practical and trustworthy for routine clinical use.

\subsection{Categorization and Comparison}
Once GPT-4 has generated and refined a set of labels based on its pre-existing knowledge base or potentially through a collaborative "human-in-the-loop" approach, the subsequent phase involves applying these labels to categorize new medical texts. This process is important for evaluating the effectiveness and applicability of the labels in real-world medical settings. The categorization of medical texts using AI-generated labels represents a significant shift from traditional text analysis methods, switching towards a more refined understanding of medical documents.

The primary goal of this phase is to assess how well the labels facilitate the grouping of texts based on shared characteristics, which may include diagnostic findings, treatment recommendations, or patient management strategies. This approach allows for a more granular and semantic analysis of medical texts, beyond what is achievable through standard lexical analysis tools. The comparison of texts based on these labels can help identify patterns, similarities, and differences across a wide array of medical documents, providing valuable insights for clinical research, patient care, and healthcare administration.

\section{Experiment}
\subsection{MIMIC-CXR}
Our experiment utilizes the MIMIC-CXR dataset \cite{johnson2019mimiccxrjpg} from Beth Israel Deaconess Medical Center, which includes de-identified chest radiograph images and detailed radiology reports. We specifically used the radiology reports section, with its rich annotations and ground truth labels, to test our text comparison methodology using GPT-4. These labels, curated by medical professionals, provide a high-accuracy benchmark for evaluating GPT-4’s performance in identifying and categorizing report content.

To ensure our analysis focused on reports offering the most clinically relevant information for GPT-4 to analyze, we excluded reports labeled solely as "No Finding" by CheXpert and NegBio. This decision was based on the rationale that such reports, while indicating the absence of specific findings, often lack the detailed clinical information necessary for GPT-4 to generate insightful labels. Consequently, our dataset curation aimed at selecting reports with descriptive labels to fully utilize GPT-4’s capabilities in generating clinically significant labels, thereby optimizing the utility of our methodology for medical text analysis.

\subsection{Utilizing GPT-4 for Label Generation}
\begin{figure*}[h]
\centering
\includegraphics[width=0.8\textwidth]{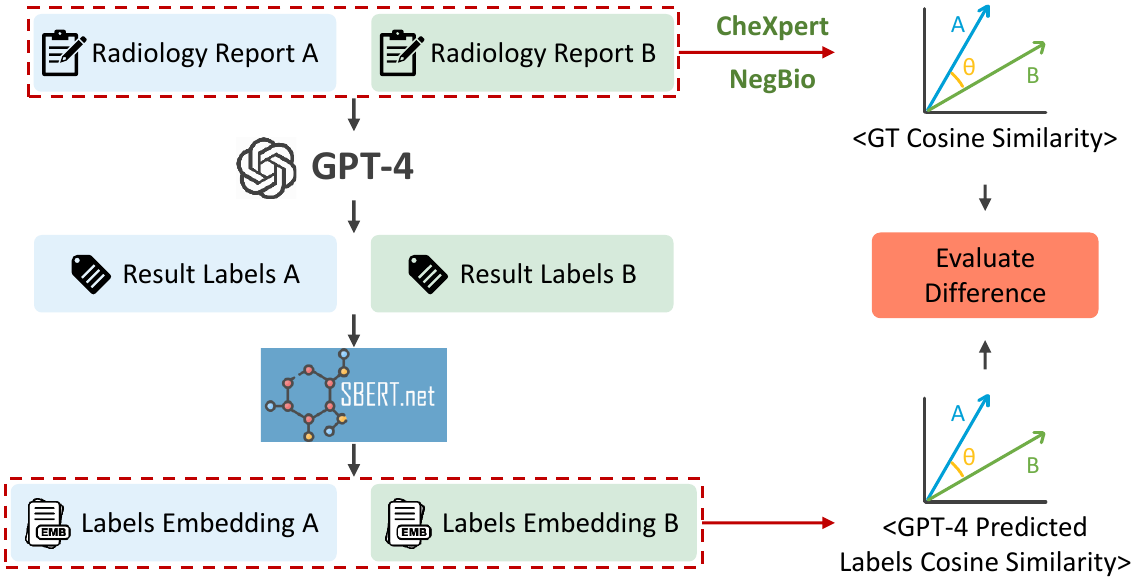}
\caption{Computational pipeline of the similarity difference calculation between two radiology reports from the MIMIC-CXR dataset used in our experimental settings.}
\label{pipeline}
\end{figure*}

Our experiment (See Figure \ref{pipeline}) began by presenting GPT-4 with four randomly chosen radiology reports from the MIMIC-CXR dataset to test its text recognition capabilities. GPT-4 accurately identified these as chest radiology reports, showcasing its potential for sophisticated text analysis. We then advanced to having GPT-4 generate tasks and corresponding generic labels for these tasks, relying solely on its pre-existing knowledge, to see how well it could classify texts within the medical domain without additional information.

After generating tasks and labels, we conducted manual reviews to select those that best matched the findings-based labeling method used in the MIMIC-CXR dataset. This ensured the generated labels closely mirrored the dataset's ground truth labels, maintaining consistency and relevance.

Next, we applied this refined methodology to new reports from the MIMIC-CXR dataset, directing GPT-4 to generate labels for each using the findings-based approach. This procedure was replicated for 500 reports, allowing us to assess GPT-4's label generation capabilities across a variety of reports. This approach provided us with a well-rounded collection of GPT-4 generated labels, confirming the model's effectiveness in producing clinically relevant labels and highlighting the consistency of our methodology.

\subsection{Obtaining Ground Truth Similarities}
In our study, we utilized a subset of 500 radiology reports from the MIMIC-CXR dataset. To facilitate a comprehensive analysis of report similarities, these reports were evenly divided into two groups, ensuring that each report from one group would be systematically compared with every report in the opposing group. This division resulted in a total of 62,500 unique comparisons, providing a robust foundation for evaluating the effectiveness of our proposed methodology in identifying similarities between radiology reports.

The core of our ground truth similarity assessment lies in the conversion of labeled data for each report into a vector representation. The labels provided by CheXpert and NegBio, which include a range of pathologies and findings such as Atelectasis, Cardiomegaly, and Pleural Effusion, were encoded into vectors comprising four possible values: 1.0 (positive mention), 0.0 (negative mention), -1.0 (uncertain or ambiguous mention), and missing. To accommodate the computational requirements of our similarity analysis, the missing values, which signify the absence of a mention of the label in the report, were replaced with '-2'. This adjustment was critical for ensuring that each aspect of the labeling—positive, negative, uncertain, and unmentioned—was distinctly represented in the vector, allowing for a distinctive comparison between reports.

We then calculated the ground truth similarity between report pairs using the cosine similarity metric, which quantifies the similarity between two vectors, with scores ranging from -1 to 1. This approach provided a measurable similarity score based on the clinical findings and diagnoses as categorized by CheXpert and NegBio labels, offering a detailed assessment of our proposed methodology's effectiveness.

\subsection{Computing Similarities for GPT-4 Generated Labels}
To evaluate the similarities between radiology reports based on labels generated by GPT-4, we employed the "all-mpnet-base-v2" model \cite{sentence-transformers_all-mpnet-base-v2}, a state-of-the-art sentence embedding framework designed to capture deep semantic meanings. This model was specifically chosen for its ability to understand and process the complex medical terminology and varied expressions found in radiology reports, translating them into a high-dimensional vector space where semantic similarities could be quantitatively assessed.

After converting the labels into normalized semantic embeddings using "all-mpnet-base-v2," we calculated the cosine similarity between label vectors from two distinct reports within our dataset, covering 62,500 comparisons. This approach quantified the semantic similarity between report label pairs, providing a detailed evaluation of the labels' relevance and coherence as generated by GPT-4.

\subsection{Evaluation of GPT-4 Generated Labels Versus Traditional Metrics}
In our comprehensive analysis, we extended the evaluation of radiology report similarities by incorporating traditional NLP metrics, including ROUGE-1, ROUGE-2, ROUGE-L, and BLEU scores, for all 62,500 unique comparisons between the reports. These metrics, widely used in the assessment of text generation and summarization tasks, provided a baseline for lexical similarity that focuses on the overlap of n-grams between the texts for ROUGE metrics and the precision of matched sequences in the BLEU score. By applying these metrics, we aimed to quantify the lexical coherence and similarity between pairs of radiology reports based on their textual content.

Following the computation of traditional similarity scores, we began the comparison procedures. Here, we calculated the difference between the scores obtained from ROUGE and BLEU metrics and the semantic similarities derived from the GPT-4 generated labels against the ground truth similarities previously established. This comparative analysis was essential for understanding how the GPT-4 generated labels' semantic depth and relevance measure against the lexical similarity benchmarks provided by ROUGE and BLEU scores.

To synthesize our findings, we calculated the mean of all differences across the comparisons for each method (ROUGE-1, ROUGE-2, ROUGE-L, BLEU, and GPT-4 generated labels) against ground truth similarities. This aggregation provided a holistic view of the variance between traditional lexical similarity metrics and the semantically driven approach offered by GPT-4 generated labels. By computing the average differences, we aimed to discern patterns and insights into the effectiveness of GPT-4's labels in capturing the true similarity between radiology reports, as compared to conventional NLP metrics.

\section{Results}
Our analysis focused on comparing predicted similarities of radiology report pairs using various methods against ground truth (GT) similarities, derived from a dataset encompassing 62,500 text pair comparisons. The comparison leveraged several metrics, including GPT\_sim (GPT-4 generated similarities), ROUGE-1 F1, ROUGE-2 F1, ROUGE-L F1, and BLEU scores, to evaluate their performance in aligning with the GT derived from CheXpert and NegBio annotations.

\begin{figure}[h]
\centering
\includegraphics[width=0.7\columnwidth]{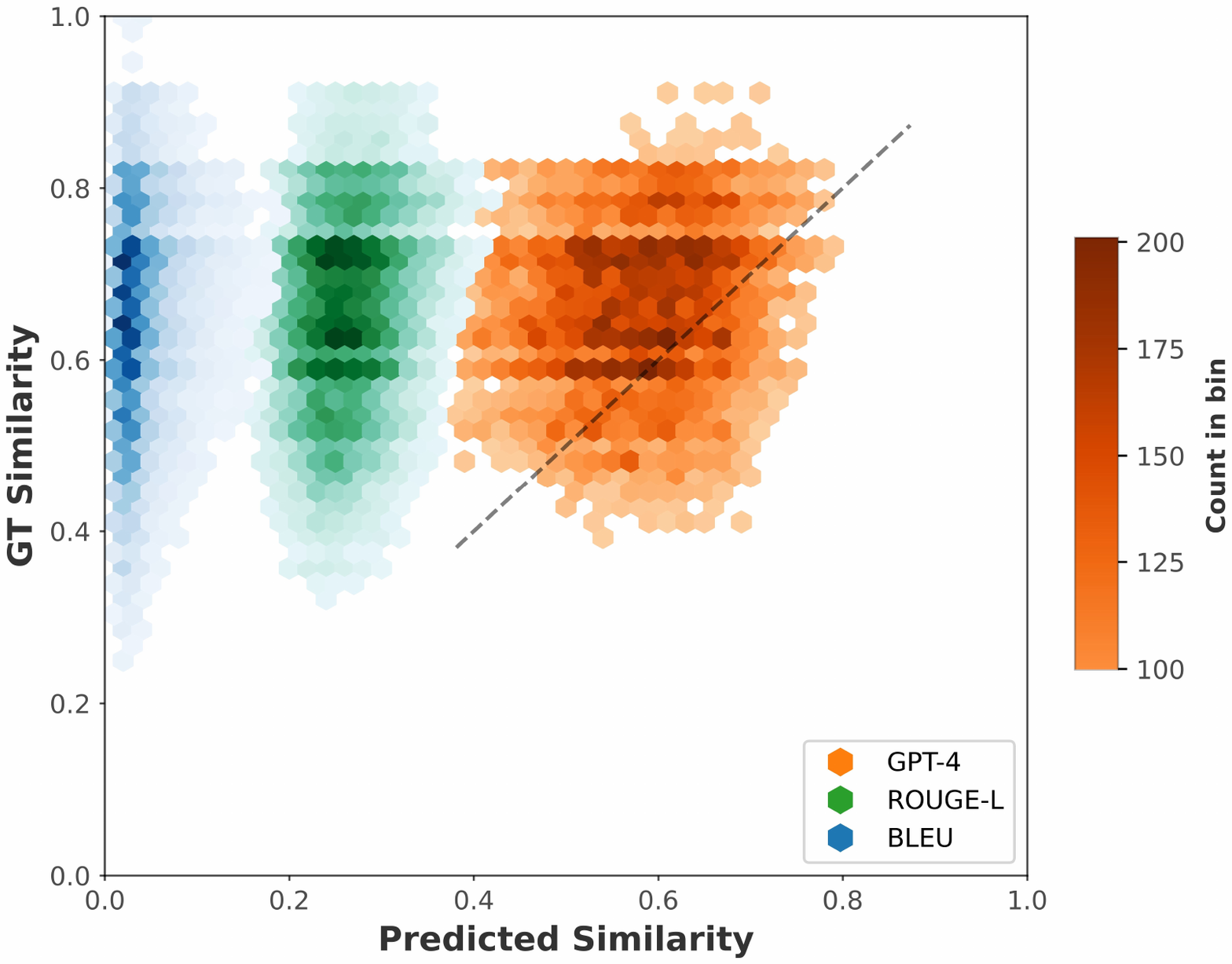}
\caption{Comparison of predicted similarities using various methods across the 62,500 text pairs. The dashed line denotes the ideal prediction-ground truth (GT) match across the 5th to 95th percentile range of GT similarities. A closer alignment to the dashed line signifies a greater correspondence with the GT. For clarity, only hexagonal bins with more than 100 observations are displayed. The results demonstrate that GPT\_sim exhibits the highest degree of alignment with the GT.}
\label{results}
\end{figure}

A visual representation of our findings is illustrated in Figure \ref{results} that maps the predicted similarities against the GT similarities. In this figure, a dashed line represents the ideal prediction-GT match, spanning from the 5th to the 95th percentile range of GT similarities. A closer proximity of the data points to this dashed line indicates a higher correspondence with the GT. For enhanced clarity and focus on significant trends, only hexagonal bins aggregating more than 100 observations are displayed.

The results from our analysis are summarized in Table \ref{tab:mean_differences}, which shows the mean similarity scores for each comparison method against both CheXpert and NegBio annotations. Notably, GPT\_sim exhibits the highest degree of alignment with the GT, achieving scores of 0.1768 and 0.1793 against CheXpert and NegBio, respectively. This contrasts with the higher mean scores observed for ROUGE and BLEU metrics, indicating a greater deviation from the GT. Specifically, ROUGE\_1\_F1, ROUGE\_2\_F1, and ROUGE\_L\_F1 scores ranged from 0.3654 to 0.5827, while BLEU scores were close to 0.6, all of which signify less correspondence with the GT compared to GPT\_sim.

These findings convey GPT\_sim's superior performance in mirroring the GT similarities, highlighting its efficacy in capturing the semantic nuances of medical texts. The marked discrepancy between GPT\_sim and traditional NLP metrics in alignment with GT shows the potential of advanced language models in enhancing the accuracy and relevance of medical document analysis. This comparison not only validates the proposed methodology's effectiveness but also illustrates the limitations of conventional metrics in reflecting the true semantic similarities within medical texts, advocating for the broader adoption of AI-driven methods in healthcare informatics.

\begin{table}[h]
\centering
\caption{Mean Differences in Similarity Scores}
\label{tab:mean_differences}
\begin{tabular}{lcc}
\hline
\textbf{Comparison} & \textbf{CheXpert} & \textbf{NegBio} \\ \hline
GPT\_sim & 0.1768 & 0.1793 \\
ROUGE\_1\_F1 & 0.3654 & 0.3714 \\
ROUGE\_2\_F1 & 0.5767 & 0.5827 \\
ROUGE\_L\_F1 & 0.3765 & 0.3825 \\
BLEU & 0.5991 & 0.5941 \\ \hline
\end{tabular}
\end{table}

\section{Limitations and Future Work}
This study, while pioneering in its approach to leveraging advanced language models for the analysis of medical texts, encounters several limitations that merit consideration. A primary limitation arises from the hypothesized 'human-in-the-loop' (HITL) component of our methodology. Despite the theoretical framework laid out for integrating medical professionals' insights into the AI-generated label evaluation process, the actual implementation of this aspect was beyond the scope of our current research. Consequently, the potential enhancements in accuracy, relevance, and clinical utility anticipated from direct human expert involvement remain speculative at this stage. The absence of medical professionals' engagement in refining and validating AI-generated labels presents a gap between the envisioned and realized capabilities of our methodology, potentially affecting the robustness of our findings and their applicability in clinical settings.

Furthermore, the scope of our research was confined to the analysis of chest x-ray radiology reports, limiting the generalizability of our findings across the vast spectrum of medical documentation. While chest x-rays constitute a significant subset of medical imaging, the medical field encompasses a wide array of data types, each with its unique linguistic and clinical characteristics. Our methodology's effectiveness in interpreting and categorizing texts from other medical domains, such as pathology reports, clinical notes, or imaging reports from other modalities, remains untested. This limitation underscores the need for further research to explore the applicability and scalability of our proposed method across different types of medical data, ensuring its relevance and utility in a broader healthcare context.

Addressing these limitations is essential for advancing the research agenda set forth by this study. Future work should aim to operationalize the HITL framework with the active participation of medical professionals, thereby empirically testing and refining the AI-generated labels' clinical validity. Additionally, extending the analysis to encompass a wider variety of medical texts will be crucial for assessing the method's versatility and effectiveness in diverse healthcare informatics applications. By confronting these challenges, subsequent research can build on our foundational work, enhancing the integration of AI in medical text analysis and contributing to the advancement of precision medicine and evidence-based clinical practices.

\section{Conclusion}
Our investigation into the semantic analysis of medical texts, particularly radiology reports from the MIMIC-CXR dataset, has indicated the potential of employing advanced language models like GPT-4, coupled with the semantic embedding capabilities of "all-mpnet-base-v2". This study's approach surpasses the conventional boundaries set by lexical comparison metrics such as ROUGE and BLEU, which, despite their utility in broad NLP applications, fall short of adequately capturing the nuanced semantic depths necessary for meaningful medical text analysis.

The methodological framework developed herein—leveraging GPT-4 for the generation of clinically relevant labels and employing semantic embeddings for comprehensive similarity assessments—has demonstrated a significant enhancement in the ability to discern semantic similarities between medical documents. Our findings, derived from extensive comparisons across a curated subset of 500 radiology reports, reveal that the labels generated by GPT-4, when analyzed through semantic embeddings, align more closely with the clinical ground truth than the traditional NLP metrics. This alignment not only indicates the superiority of our proposed method in capturing the essence of medical texts but also highlights the potential of such AI-driven approaches to revolutionize healthcare informatics by facilitating a deeper understanding of clinical documentation.

However, the study acknowledges certain limitations, including the absence of direct medical professional involvement in the label evaluation process and the focus solely on chest radiology reports. These limitations delineate avenues for future research, particularly emphasizing the need for integrating "human-in-the-loop" methodologies to enhance label accuracy and relevance, as well as expanding the analysis to encompass a broader spectrum of medical documents.

In conclusion, the outcomes of this study advocate for a shift in the application of AI within medical text analysis, illustrating the benefits of moving towards more semantically aware models. By combining the computational strengths of GPT-4 with the insights of human expertise, our research contributes an important step forward in the advancement of precision medicine and evidence-based clinical practices. As we continue to refine these methodologies and explore their application across wider domains of medical data, the potential to greatly improve patient care through enhanced data analysis and interpretation becomes increasingly apparent, marking a new chapter in the intersection of AI and healthcare.

\bibliographystyle{unsrt}  
\bibliography{references}  

\begin{thebibliography}{10}

\bibitem{lin-2004-rouge}
Chin-Yew Lin.
\newblock {ROUGE}: A package for automatic evaluation of summaries.
\newblock In {\em Text Summarization Branches Out}, pages 74--81, Barcelona, Spain, July 2004. Association for Computational Linguistics.

\bibitem{papineni-etal-2002-bleu}
Kishore Papineni, Salim Roukos, Todd Ward, and Wei-Jing Zhu.
\newblock {B}leu: a method for automatic evaluation of machine translation.
\newblock In Pierre Isabelle, Eugene Charniak, and Dekang Lin, editors, {\em Proceedings of the 40th Annual Meeting of the Association for Computational Linguistics}, pages 311--318, Philadelphia, Pennsylvania, USA, July 2002. Association for Computational Linguistics.

\bibitem{liu2023geval}
Yang Liu, Dan Iter, Yichong Xu, Shuohang Wang, Ruochen Xu, and Chenguang Zhu.
\newblock G-eval: Nlg evaluation using gpt-4 with better human alignment, 2023.

\bibitem{naismith2023automated}
Ben Naismith, Phoebe Mulcaire, and Jill Burstein.
\newblock Automated evaluation of written discourse coherence using gpt-4.
\newblock In {\em Proceedings of the 18th Workshop on Innovative Use of NLP for Building Educational Applications (BEA 2023)}, pages 394--403, 2023.

\bibitem{fu2023gptscore}
Jinlan Fu, See-Kiong Ng, Zhengbao Jiang, and Pengfei Liu.
\newblock Gptscore: Evaluate as you desire, 2023.

\bibitem{brown2020language}
Tom~B. Brown, Benjamin Mann, Nick Ryder, Melanie Subbiah, Jared Kaplan, Prafulla Dhariwal, Arvind Neelakantan, Pranav Shyam, Girish Sastry, Amanda Askell, Sandhini Agarwal, Ariel Herbert-Voss, Gretchen Krueger, Tom Henighan, Rewon Child, Aditya Ramesh, Daniel~M. Ziegler, Jeffrey Wu, Clemens Winter, Christopher Hesse, Mark Chen, Eric Sigler, Mateusz Litwin, Scott Gray, Benjamin Chess, Jack Clark, Christopher Berner, Sam McCandlish, Alec Radford, Ilya Sutskever, and Dario Amodei.
\newblock Language models are few-shot learners, 2020.

\bibitem{devlin2019bert}
Jacob Devlin, Ming-Wei Chang, Kenton Lee, and Kristina Toutanova.
\newblock Bert: Pre-training of deep bidirectional transformers for language understanding, 2019.

\bibitem{geminiteam2023gemini}
Gemini Team.
\newblock Gemini: A family of highly capable multimodal models, 2023.

\bibitem{arrieta2019explainable}
Alejandro~Barredo Arrieta, Natalia Díaz-Rodríguez, Javier~Del Ser, Adrien Bennetot, Siham Tabik, Alberto Barbado, Salvador García, Sergio Gil-López, Daniel Molina, Richard Benjamins, Raja Chatila, and Francisco Herrera.
\newblock Explainable artificial intelligence (xai): Concepts, taxonomies, opportunities and challenges toward responsible ai, 2019.

\bibitem{holzinger2016interactive}
Andreas Holzinger.
\newblock Interactive machine learning for health informatics: when do we need the human-in-the-loop?
\newblock {\em Brain Informatics}, 3(2):119--131, 2016.

\bibitem{reiter2018structured}
Ehud Reiter.
\newblock A structured review of the validity of bleu.
\newblock {\em Computational Linguistics}, 44(3):393--401, 2018.

\bibitem{gulden2019extractive}
Christian Gulden, Melanie Kirchner, Christina Sch{\"u}ttler, Marc Hinderer, Marvin Kampf, Hans-Ulrich Prokosch, and Dennis Toddenroth.
\newblock Extractive summarization of clinical trial descriptions.
\newblock {\em International journal of medical informatics}, 129:114--121, 2019.

\bibitem{banerjee2005meteor}
Satanjeev Banerjee and Alon Lavie.
\newblock Meteor: An automatic metric for mt evaluation with improved correlation with human judgments.
\newblock In {\em Proceedings of the acl workshop on intrinsic and extrinsic evaluation measures for machine translation and/or summarization}, pages 65--72, 2005.

\bibitem{puppala2015meteor}
Mamta Puppala, Tiancheng He, Shenyi Chen, Richard Ogunti, Xiaohui Yu, Fuhai Li, Robert Jackson, and Stephen~TC Wong.
\newblock Meteor: an enterprise health informatics environment to support evidence-based medicine.
\newblock {\em IEEE Transactions on Biomedical Engineering}, 62(12):2776--2786, 2015.

\bibitem{soualmia2012matching}
Lina~F Soualmia, Elise Prieur-Gaston, Zied Moalla, Thierry Lecroq, and St{\'e}fan~J Darmoni.
\newblock Matching health information seekers' queries to medical terms.
\newblock {\em BMC bioinformatics}, 13(14):1--15, 2012.

\bibitem{rahutomo2012semantic}
Faisal Rahutomo, Teruaki Kitasuka, and Masayoshi Aritsugi.
\newblock Semantic cosine similarity.
\newblock In {\em The 7th international student conference on advanced science and technology ICAST}, volume~4, page~1, 2012.

\bibitem{huang2016supervised}
Gao Huang, Chuan Guo, Matt~J Kusner, Yu~Sun, Fei Sha, and Kilian~Q Weinberger.
\newblock Supervised word mover's distance.
\newblock {\em Advances in neural information processing systems}, 29, 2016.

\bibitem{ye2015improved}
Jun Ye.
\newblock Improved cosine similarity measures of simplified neutrosophic sets for medical diagnoses.
\newblock {\em Artificial intelligence in medicine}, 63(3):171--179, 2015.

\bibitem{xia2015learning}
Peipei Xia, Li~Zhang, and Fanzhang Li.
\newblock Learning similarity with cosine similarity ensemble.
\newblock {\em Information sciences}, 307:39--52, 2015.

\bibitem{kusner2015word}
Matt Kusner, Yu~Sun, Nicholas Kolkin, and Kilian Weinberger.
\newblock From word embeddings to document distances.
\newblock In {\em International conference on machine learning}, pages 957--966. PMLR, 2015.

\bibitem{gong2023evaluating}
Xinyu Gong, Jason Holmes, Yiwei Li, Zhengliang Liu, Qi~Gan, Zihao Wu, Jianli Zhang, Yusong Zou, Yuxi Teng, Tian Jiang, et~al.
\newblock Evaluating the potential of leading large language models in reasoning biology questions.
\newblock {\em arXiv preprint arXiv:2311.07582}, 2023.

\bibitem{shi2023mededit}
Yucheng Shi, Shaochen Xu, Zhengliang Liu, Tianming Liu, Xiang Li, and Ninghao Liu.
\newblock Mededit: Model editing for medical question answering with external knowledge bases.
\newblock {\em arXiv preprint arXiv:2309.16035}, 2023.

\bibitem{kim2023medivista}
Sekeun Kim, Kyungsang Kim, Jiang Hu, Cheng Chen, Zhiliang Lyu, Ren Hui, Sunghwan Kim, Zhengliang Liu, Aoxiao Zhong, Xiang Li, et~al.
\newblock Medivista-sam: Zero-shot medical video analysis with spatio-temporal sam adaptation.
\newblock {\em arXiv preprint arXiv:2309.13539}, 2023.

\bibitem{liu2023radonc}
Zhengliang Liu, Peilong Wang, Yiwei Li, Jason Holmes, Peng Shu, Lian Zhang, Chenbin Liu, Ninghao Liu, Dajiang Zhu, Xiang Li, et~al.
\newblock Radonc-gpt: A large language model for radiation oncology.
\newblock {\em arXiv preprint arXiv:2309.10160}, 2023.

\bibitem{liu2023artificial}
Chenbin Liu, Zhengliang Liu, Jason Holmes, Lu~Zhang, Lian Zhang, Yuzhen Ding, Peng Shu, Zihao Wu, Haixing Dai, Yiwei Li, et~al.
\newblock Artificial general intelligence for radiation oncology.
\newblock {\em Meta-Radiology}, page 100045, 2023.

\bibitem{liu2023holistic}
Zhengliang Liu, Hanqi Jiang, Tianyang Zhong, Zihao Wu, Chong Ma, Yiwei Li, Xiaowei Yu, Yutong Zhang, Yi~Pan, Peng Shu, et~al.
\newblock Holistic evaluation of gpt-4v for biomedical imaging.
\newblock {\em arXiv preprint arXiv:2312.05256}, 2023.

\bibitem{zhao2023ophtha}
Huan Zhao, Qian Ling, Yi~Pan, Tianyang Zhong, Jin-Yu Hu, Junjie Yao, Fengqian Xiao, Zhenxiang Xiao, Yutong Zhang, San-Hua Xu, et~al.
\newblock Ophtha-llama2: A large language model for ophthalmology.
\newblock {\em arXiv preprint arXiv:2312.04906}, 2023.

\bibitem{zhong2023chatradio}
Tianyang Zhong, Wei Zhao, Yutong Zhang, Yi~Pan, Peixin Dong, Zuowei Jiang, Xiaoyan Kui, Youlan Shang, Li~Yang, Yaonai Wei, et~al.
\newblock Chatradio-valuer: A chat large language model for generalizable radiology report generation based on multi-institution and multi-system data.
\newblock {\em arXiv preprint arXiv:2310.05242}, 2023.

\bibitem{avati2018improving}
Anand Avati, Kenneth Jung, Stephanie Harman, Lance Downing, Andrew Ng, and Nigam~H Shah.
\newblock Improving palliative care with deep learning.
\newblock {\em BMC medical informatics and decision making}, 18(4):55--64, 2018.

\bibitem{caruana2015intelligible}
Rich Caruana, Yin Lou, Johannes Gehrke, Paul Koch, Marc Sturm, and Noemie Elhadad.
\newblock Intelligible models for healthcare: Predicting pneumonia risk and hospital 30-day readmission.
\newblock In {\em Proceedings of the 21th ACM SIGKDD international conference on knowledge discovery and data mining}, pages 1721--1730, 2015.

\bibitem{johnson2019mimiccxrjpg}
Alistair E.~W. Johnson, Tom~J. Pollard, Nathaniel~R. Greenbaum, Matthew~P. Lungren, Chih ying Deng, Yifan Peng, Zhiyong Lu, Roger~G. Mark, Seth~J. Berkowitz, and Steven Horng.
\newblock Mimic-cxr-jpg, a large publicly available database of labeled chest radiographs, 2019.

\bibitem{sentence-transformers_all-mpnet-base-v2}
HuggingFace.
\newblock all-mpnet-base-v2.
\newblock \url{https://huggingface.co/sentence-transformers/all-mpnet-base-v2}.
\newblock Accessed: 2024-02-15.

\end{thebibliography}

\end{document}